\documentclass[10pt,letterpaper,twocolumn]{article}
\usepackage{array}
\usepackage{multirow}
\usepackage{amsmath}
\usepackage{amssymb}
\usepackage{graphicx}
\usepackage{float}
\usepackage{epstopdf}
\usepackage{url}

\newif\ifpdf\ifx\pdfoutput\undefined\pdffalse\else\pdfoutput=1\pdftrue\fi

\usepackage{graphicx}

\usepackage{booktabs}

\makeatletter

\pdfpageheight\paperheight
\pdfpagewidth\paperwidth

\usepackage{cvpr}
\usepackage{times}
\usepackage{epsfig}

\usepackage{algorithmicx}
\usepackage[ruled]{algorithm}
\usepackage[noend]{algpseudocode}
\usepackage{setspace}

 \cvprfinalcopy 

\def\cvprPaperID{1794} 

\makeatother

\begin{document}

\title{Compositional Memory for Visual Question Answering}

\author{Aiwen Jiang$^{1,2}$ $~~~~~~~~~$ $~~~~~~~~~$Fang Wang$^{2}$  $~~~~~~~~~$ $~~~~~~~~~$ Fatih Porikli$^{2}$ $~~~~~~~~~$ $~~~~~~~~~$ Yi Li$^*~^{2,3}$\\
\\
$^{1}$Jiangxi Normal University$~~~~~~~~~$  $^{2}$NICTA and ANU$~~~~~~~~~$ $~~~~~~~~~$ $^3$Toyota Research Institute North America\\
$^{1}$aiwen.jiang@nicta.com.au $~~~~$ $^2$\{fang.wang, fatih.porikli\}@nicta.com.au $~~~~$ $^{3}$yi.li@tema.toyota.com}

\maketitle

\begin{abstract}
Visual Question Answering (VQA) emerges as one of the most fascinating topics in computer vision recently. Many state of the art methods naively use holistic visual features with language features into a Long Short-Term Memory (LSTM) module, neglecting the sophisticated interaction between them. This coarse modeling also blocks the possibilities of exploring finer-grained local features that contribute to the question answering dynamically over time. 

This paper addresses this fundamental problem by directly modeling the temporal dynamics between language and all possible local image patches. When traversing the question words sequentially, our end-to-end approach explicitly fuses the features associated to the words and the ones available at multiple local patches in an attention mechanism, and further combines the fused information to generate dynamic messages, which we call episode. We then feed the episodes to a standard question answering module together with the contextual visual information and linguistic information. Motivated by recent practices in deep learning, we use auxiliary loss functions during training to improve the performance. Our experiments on two latest public datasets suggest that our method has a superior performance. Notably, on the DAQUAR dataset we advanced the state of the art by 6$\%$, and we also evaluated our approach on the most recent MSCOCO-VQA dataset.
\end{abstract}

\section{Introduction}

Given an image and a question, the goal of Visual Question Answering (VQA) is to directly infer the answer(s) automatically from the image. This is undoubtedly one of the most interesting, and arguably one of the most challenging, topics in computer vision in recent times.

Almost all state-of-the-art methods use predominantly holistic visual features in their systems. For example, Malinowski \textit{et al.} \cite{Malinowski-ICCV15} used the concatenation of linguistic feature and visual feature extracted by a Convolutional Neural Network (CNN), and Ren \textit{et al.} \cite{ren2015imageqa} considered the visual feature as the first word to initialize the sequential learning.

While the use of holistic approach is straightforward and convenient, it is, however, debatably problematic. For example, in the VQA problems many answers are directly related to the contents of some image regions. Therefore, it is dubious if the holistic features are rich enough to provide the information only available at regions. Also, it may hinder the exploration of finer-grained local features for VQA.

In this paper we propose a Compositional Memory for an end-to-end training framework. Our approach takes the advantage of the recent progresses in image captioning \cite{LRCN,karpathy2014deep}, natural language processing \cite{DBLP:journals/corr/KumarISBEPOGS15}, and computer vision to advance the study of the VQA. Our goal is to fuse local visual features and the linguistic information over time, in a Long Short-Term Memory (LSTM) based framework. The fused information, which we call ``episodes'', characterizes the interaction and dynamics between vision and language.

Explicitly addressing the interaction between question words and local visual features has a number of advantages. To begin with, regions provide rich info towards capturing the dynamics in question. Intuitively, parts of an image serve as ``candidates'' that may have varying importance at different time when parsing a question sentence. Recent study of image captioning \cite{icml2015_xuc15}, a closely related research topic, suggests that visual attention mechanism is very important in generating good descriptions. Obviously, this idea will also improve the accuracy in question answering.

Going deeper, this candidacy is closely related to the concept of semantic ``facts'' in reasoning. For example, one often begins to dynamically search useful local visual evidences at the same time when (s)he reads words. The use of facts has been explored in the natural language processing recently \cite{DBLP:journals/corr/KumarISBEPOGS15}, but this useful concept cannot be explored without local visual information in computer vision.

While the definition of ``visual facts'' is still elusive, we can approach the problem through modeling interactions between vision and language. This ``sensory interaction'' plays a phenomenal role in the information processing and reasoning. It has a significant meaning in memory study as well. Eichenbaum and Cohen argued that part of the human memory needs to be modeled as a form of relationship between spatial, sensory and temporal information \cite{cohen1995memory}.



Specifically, our method traverses the words in a question sequentially, and explicitly fuses the linguistic and the ones available at local patches to episodes. An attention mechanism  is used to re-weight the importance of the regions \cite{icml2015_xuc15}.
The fused information is fed to a dynamic network to generate episodes. 
We then feed the episodes to a standard question answering module together with the contextual visual information and linguistic information.

The use of local features inevitably leads to the quest about region selection.
In principle, the regions can be 1) patches generated by object proposals, such as those obtained by edgebox \cite{edgebox2014} and faster-RCNN \cite{ren2015fasterRCNN}, and 2) overlapping patches that cover most important contents in image. In this paper, we choose the latter and use the features of the last convolutional layer in the CNNs.

Our experiments on two latest public datasets suggest that our method outperforms the other state of the art methods. We tested on the  DAQUAR \cite{Malinowski_NIPS2014} and MSCOCO-VQA \cite{MSCOCOVQA2015}. Notably, on the DAQUAR dataset we advanced the state of the art by 6$\%$.
We also compared a few variants of our method and demonstrated the usefulness of the Compositional Memory for VQA.
We further verified our idea on the latest MSCOCO-VQA dataset.

The main contributions of the paper are: 
\begin{itemize}
	\item We present an end-to-end approach that explores the local fine grained visual information for VQA tasks, 
    \item We develop a Compositional Memory that explicitly models the interactions of vision and language,
    \item Our method has a superior performance and it outperforms the state of the art methods.
\end{itemize}

\section{Related Work}
\subparagraph{CNN, RNN, and LSTM}
Recently, deep learning has achieved great success on many computer vision tasks. For example, CNN has set records on standard object recognition benchmarks \cite{Krizhevsky2012}. With a deep structure, CNN can effectively learn complicated mappings from raw images to the target, which requires less domain knowledge compared to handcrafted features and shallow learning frameworks. 

Recurrent Neural Networks (RNN) have been used for modeling temporal sequences and gained attention in speech recognition \cite{RNN4Speech2913}, machine translation \cite{RNN4MachineTranlation2015}, image captioning \cite{LRCN, karpathy2014deep}. The recurrent connections are feedback loops in the unfolded network, and because of these connections, RNNs are suitable for  modeling time series with strong nonlinear dynamics and long time correlations. The traditional RNN is hard to train due to the vanishing gradient problem, \textit{i.e.} the weight updates computed via error backpropagation through time may become very small.

Long Short Term Memory model \cite{hochreiter1997long} has been proposed as a solution to overcome these problems. The LSTM architecture uses memory cells with gated access to store and output information, which alleviates the vanishing gradient problem in backpropagation over multiple time steps. Specifically, in addition to the hidden state, the LSTM also includes an input gate, a forget gate, an output gate, and the memory cell.
In this architecture, input gate and forget gate are sigmoidal gating functions, and these two terms learn to control the portions of the current input and the previous memory that the LSTM takes into consideration for overwriting the previous state. Meanwhile, the output gate controls how much of the memory should be transferred to the hidden state. These mechanisms allow LSTM networks to learn temporal dynamics. 


\subparagraph{Language and vision} The effort of combining language and vision attracts a lot of attention recently. 
Image captioning and VQA are two most intriguing problems.

Question answering (QA) is a classical problem in natural language processing \cite{DBLP:journals/corr/KumarISBEPOGS15}. When images are involved, the goal of VQA is to infer the answer of a question directly from the image \cite{MSCOCOVQA2015}. Multiple questions and answers can be associated to the same image during training.

It has been shown that VQA can borrow the idea from image captioning.
Being a related area, image captioning also uses RNN for sentence generation \cite{LRCN}.
Attention mechanism is recently adopted in image captioning and proves to be a useful component \cite{icml2015_xuc15}.

\begin{figure}[th]
  \centering
    \includegraphics[width=0.85\columnwidth]{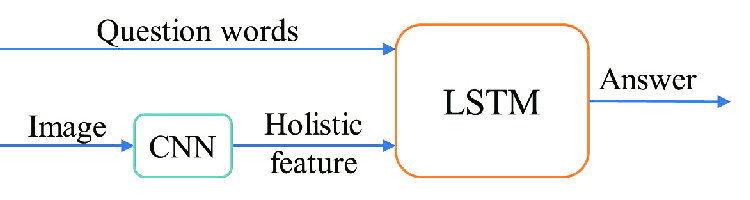}
  \caption{Classical LSTM-VQA model.}
  \label{fig:basicVQALSTM}
\end{figure}

\subparagraph{LSTM for VQA} Because a VQA system needs to process language and visual information simultaneously, most recent work adopted the LSTM in their approaches.
A typical LSTM-VQA uses holistic image features extracted by CNNs as ``visual words'', as shown in Figure~\ref{fig:basicVQALSTM}. 
The visual word features are used either as the first or at the end of question sequence \cite{ren2015imageqa} or they are concatenated together with question word vectors into a LSTM \cite{Malinowski-ICCV15,gao2015you} .

This LSTM-VQA framework is straightforward and useful.
However, treating the feature interaction as feature vector concatenation lacks the capability that explicitly extracts finer-grained information. As we discussed before, details of facts may be neglected if global visual features are used. This leads to the quest for more effective information fusion model for language and image in VQA problems.

\section{Our Approach}

We present our approach in this section. First, we present our model overview. Then, we discuss the technical details and explain the training process.

\subsection{Our End-to-End VQA model}

\begin{figure}[th]
  \centering
    \includegraphics[width=0.9\columnwidth]{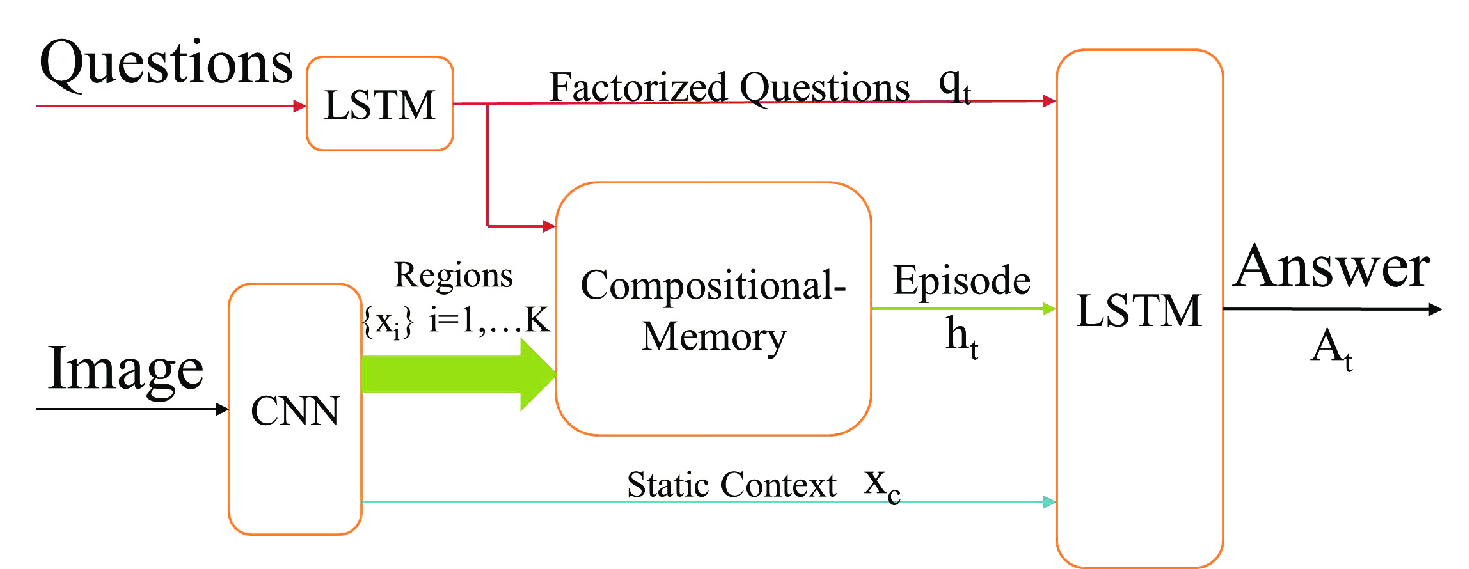}
  \caption{Proposed VQA network. Please see Figure~\ref{fig:compositelstm_unit} for the description of the Compositional Memory block}
  \label{fig:DynamicMemoryNetwork}
\end{figure}

Compared to the basic LSTM approach for VQA in Figure~\ref{fig:basicVQALSTM}, We made two major improvements of the model.
The diagram of the network is shown in Figure~\ref{fig:DynamicMemoryNetwork}.

The first, and the most important addition, is a Compositional Memory.
It reasons over those input features to produce an image-level representation for answer module. 
It reflects an experience over image contents.

The second addition is the LSTM module for parsing (factorizing) the question.
It provides input for both the Compositional Memory and the question answering LSTM.
In the experiment we will show the importance of this module for the VQA tasks.

In part, this implementation is aligned with the findings in cognitive neuroscience.
It is well known that semantic (\textit{e.g.}, visual features and classifiers) and episodic memory (\textit{e.g.}, temporal questioning sentence) together make up the declarative memory of human beings, and the interactions among them become the key in representation and reasoning \cite{tulving1972episodic}.
Our model captures this interaction naturally.

\subsubsection{Compositional Memory}
\begin{figure}[th]
  \centering
    \includegraphics[width=0.9\columnwidth]{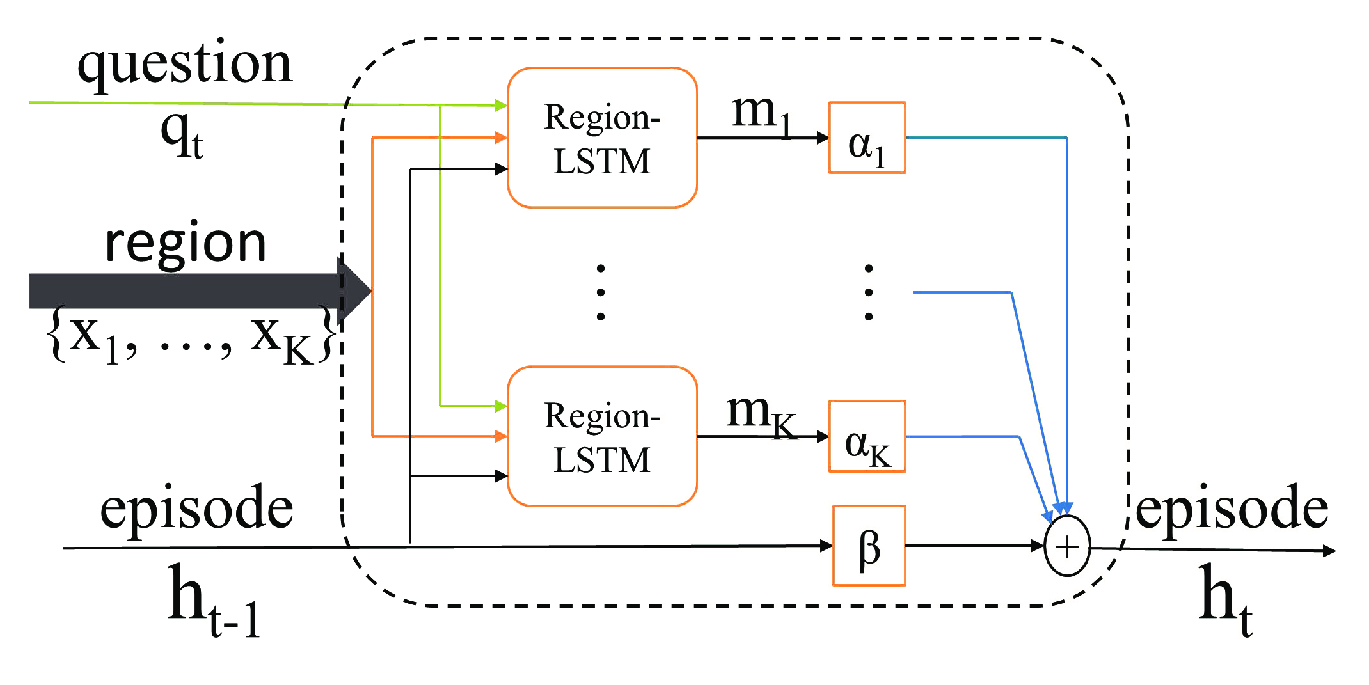}
  \caption{Diagram of Compositional Memory.}
  \label{fig:compositelstm_unit}
\end{figure}
\begin{figure*}[ht]
  \centering
    \includegraphics[width=0.98\textwidth]{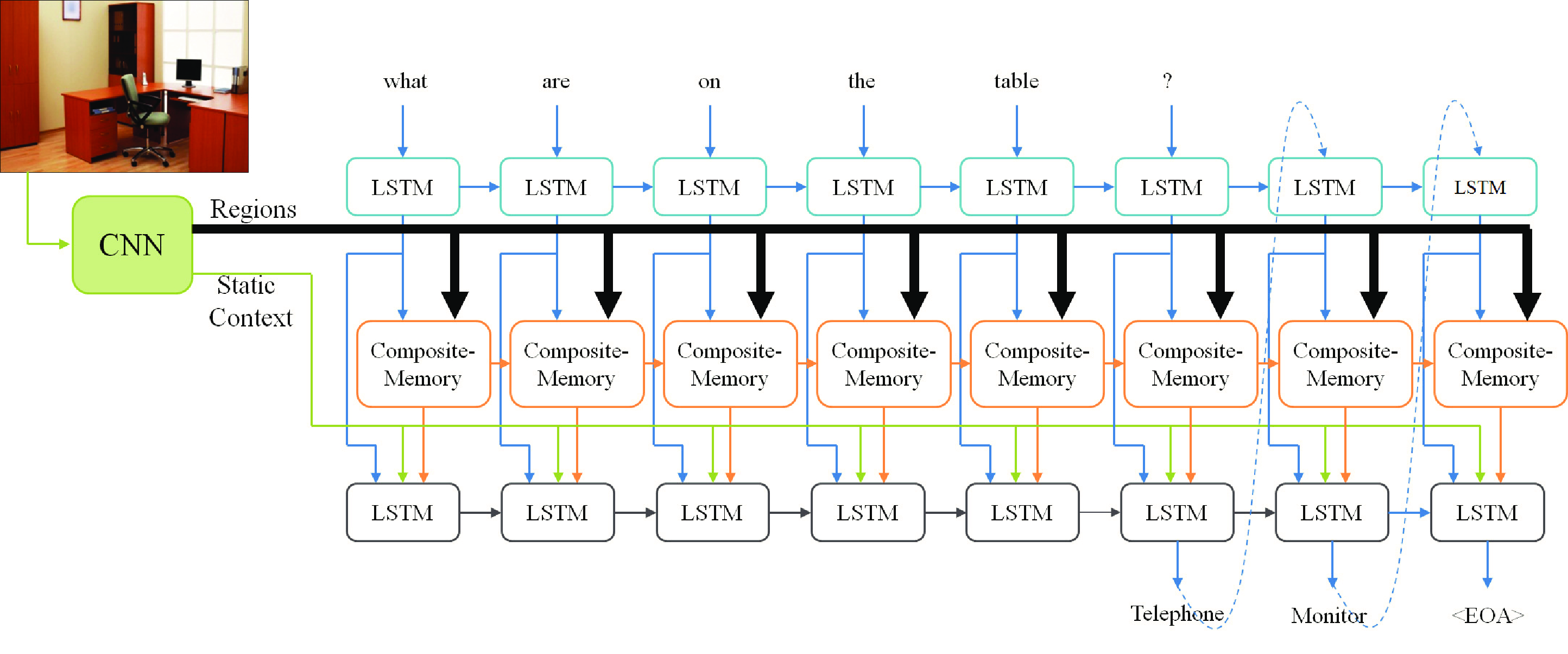}
  \caption{Unfolding our end-to-end VQA model. CM: Compositional Memory.}
  \label{fig:unfold_diagram}
\end{figure*}

We present the details of our Compositional Memory in this section.
This unit consists of a number of Region-LSTM and $\alpha$ gates (Figure~\ref{fig:compositelstm_unit}).
All these Region-LSTM share the same set of parameters, and their results are combined to generate an episode at each time step.

The region-LSTMs are mainly in charge of processing input image region contents in parallel. It dynamically generates language-embedded visual information for each region, conditioned on previous episodes, local visual feature, and current question word.
The state, gates and cells of of each Region-LSTM are updated as follows:
\begin{align}
 & i_t ^k = \sigma(W_{qi}q_t + W_{hi}h_{t-1} + W_{xi} X_k + b_i) \label{eq:lstm-first}\\
 & f_t ^k= \sigma(W_{qf}q_t + W_{hf}h_{t-1} + W_{xf} X_k + b_f) \\
 & o_t ^k= \sigma(W_{qo}q_t + W_{ho}h_{t-1} + W_{xo} X_k + b_o) \\
 & g_t ^k= \tanh(W_{qg}q_t + W_{hg}h_{t-1} + W_{xg} X_k + b_g) \\
 & c_t ^k= f_t ^k \odot c_{t-1} ^k + i_t ^k \odot g_t ^k\\
 & m_t ^k= o_t ^k \odot \tanh(c_t ^k)
 \label{eq:lstm-last}
\end{align}
where the hidden state $h_t$ denotes an episode at every time step, $q_t$ is the language information (\textit{e.g.,} features generated by \textit{word2vec}), and $X_k$ is the $k^{th}$ regional CNN features. 
$m_t$ is the output of region-LSTM $t$.
Please note that the superscripts are omitted in the above notations for simplicity.

In Region-LSTM\footnote{Please note that the diagram is not drawn for the purpose of simplicity.}, $c_t$ denotes the memory cell, $g_t$ denotes an input modulation gate. 
$i_t$ and $f_t$ are input and forget gates, which control the portions of the current input and the previous memory that LSTM takes into consideration. 
$o_t$ is output gate that determines how much of the memory to transfer to the hidden state.
$\odot$ is the element-wise multiplication operation, and $\sigma()$ and $tanh()$ denote the sigmoidal and tanh operations, respectively. 
These mechanisms allow LSTM to learn long-term temporal dynamics.

The implementation of Region-LSTM is similar to that of traditional LSTM. However, the important differences lie in the parallel strategy that all parameters ($W_{q*}, W_{h*}, W_{x*}, b_{*}$) are shared across different regions. Please note that each Region-LSTM has its own gate and cell state, respectively.

The $\alpha$ gate is also conditioned on previous episode $h_{t-1}$, region feature $X_k$, and current input language feature $q_t$. It returns a single scalar for each region.
This gate is mainly used for weighted combination of region messages for generating episodes, which are dynamically pooled into image-level information. 
Similar to $W_{zq}, W_{zh}, W_{zx}, b_z, W_\alpha, b_\alpha$, the parameters of $\alpha$ gate, are also shared across regions. At every time step $t$, $\alpha$ gate dynamically generates values $\alpha _k ^t$ for $k^{th}$ region.
\begin{equation}
\begin{array}{l}
{z_k ^t} = \tanh \left( {{W_{zq}}{q_t} + {W_{zh}}{h_{t - 1}} + {W_{zx}}{x_k} + {b_z}} \right)\\
{\alpha _k ^t} = \sigma \left( {{W_\alpha }{z_k ^t} + {b_\alpha }} \right)
\end{array}
\end{equation}

In order to summarize region-level information to an image-level feature, we employ a modified pooling mechanism of gated recurrent style \cite{GRU2014}. The episode $h_t$ acts as dynamic feature as well as hidden state of Compositional Memory unit. It is updated to renew the input information of Region-LSTMs together with language input at every time step.
\begin{equation}
\label{eq:hidden-state}
\begin{array}{l}
 \beta  = 1 - \frac{1}{K}\sum\limits_k {{\alpha _k}}\\
 {h_t} =  \beta {h_{t - 1}} + \frac{1}{K} \sum\limits_k {{\alpha _k ^t}m_t^k}
\end{array}
\end{equation}
where $K$ is the total number of regions.

One of the advantages of Compositional Memory is that it goes beyond traditional static image feature. The unit incorporates both the merits of LSTM and attention mechanism, and thus it is suitable to generate dynamic episodic messages for visual question answering applications.

\subsubsection{Language LSTM} 

We use a Language LSTM to process linguistic inputs.
We consider this representation as ``factorization of questions'', which captures the recurrent relations of question word sequence, and stores semantic memory information about the questions. This strategy for processing language has also been proven to be  important in image captioning \cite{LRCN}. 

\subsubsection{Other components}

\subparagraph{Answer generation} The answer generation module takes dynamic episodes, together with language and static visual context generated by CNNs to generate the answers in a LSTM framework.

\subparagraph{CNN} Our approach is compatible with all the major CNN network, such as AlexNet \cite{AlexNet_NIPS2012} and GoogLeNet \cite{googlenet2015}.
In each CNN, we use the last convolution layer as region selections.
For example, GoogleNet, we use the $1024 \times 7 \times 7$ feature map from the inception\_5b/output. This means that Compositional Memory operates on 49 regions, each of which is represented by 1024-dim feature vector.
Following the current practice, we used the output of the last CNN layer as visual context.

\subsection{Training}

Generally, there are three types of the VQA tasks: 1) Single Word, 2) Multiple Words (or called ``Open-ended''), and 3) Multiple Choice. The single word category restricts the answer to have only one single word. Multiple Words VQA is an open-ended task, where the length of answer is not limited and the VQA system needs to sequentially generate possible answer words until generating an answer-ending mark. The multiple choice task refers to select most probable answer from a set of answer candidates.

Among these three, the ``Open-ended'' VQA task is the most difficult one, thus we chose this category to demonstrate the performance of our approach.
In this section, we briefly present the training procedure, the loss function, and other implementation details.

\subsubsection{Protocol}
During the training of our VQA system, CNN and LSTMs are jointly learned in an end-to-end way. The unfolded network of our proposed model is shown in Figure~\ref{fig:unfold_diagram}.

In this Open-ended category of VQA, questions may have multiple word answers. 
We consequently decompose the problem to predict a set of answer words $A = \left\{ {{a_1},{a_2},...,{a_M}} \right\}$, where $a_i$ are words from a finite vocabulary $\Omega '$ and M is the number of answer words for a given question and image. To deal with open-ended VQA task, we add an extra token $\langle EOA \rangle$ into the vocabulary $\Omega  = \Omega ' \cup \{  \langle EOA \rangle \} $. The $\langle EOA \rangle$ indicates the end of the answer sequence. Therefore, we formulate the prediction procedure recursively as:
\begin{align}
&{\widehat a_t} = \arg \max p(a|X,q,{\widehat A_{t - 1}};\vartheta )
\end{align}
where ${\widehat A_{t - 1}}$ is the set of previously predicted answer words, with ${\widehat A_{0}} = \{\}$ at start. The prediction procedure is terminated when ${\widehat a_t} = \langle EOA \rangle$.

As shown in Figure~\ref{fig:DynamicMemoryNetwork} and Figure~\ref{fig:unfold_diagram}, we feed the VQA system with a question as a sequence of words, \textit{i.e.} $q = [{q_1},{q_2},...,{q_{n - 1}},\left[\kern-0.15em\left[ ? \right]\kern-0.15em\right]]$, where $\left[\kern-0.15em\left[ ? \right]\kern-0.15em\right]$ encodes the end of question. In the training phase, we augment the question word sequence with the corresponding ground truth answer sequence $a$, \textit{i.e.} $\widehat q: = [q,a]$. During the test phase, at $t^{th}$ time step we augment question $q$, with previously predicted answer words ${\widehat q_t}: = [q,{\widehat a_{1,...,t - 1}}]$.

\subsubsection{Loss function}

All the parameters are jointly learned with cross-entropy loss. The output predictions that occur before the ending question mark $\left[\kern-0.15em\left[ ? \right]\kern-0.15em\right]$ are excluded from the loss computation, so that the model is solely penalized based on the predicted answer words.
  
Motivated by the recent success of GoogLeNet, We adopt a multi-task training strategy for learning the parameters of our network.
Specifically, in additional to the question answering LSTM, we add a ``Language Only'' loss layer on the Language LSTM, and an ``Episode Only'' loss layer on the Compositional Memory.
These two auxiliary loss functions are added during training to improve the performance, and they are removed during testing.

\subsubsection{Implementation} 

We implemented our end-to-end VQA network using Caffe\footnote{\url{http://caffe.berkeleyvision.org/}}. The CNN models are pre-trained, and then fine-tuned in our recurrent network training. The source code of our implementation will be available in public.

\section{Experiments}\label{sec:exp}
We test our approach on two large data sets, namely, DAQUAR \cite{Malinowski_NIPS2014} and MSCOCO-VQA\footnote{\url{http://www.visualqa.org}} \cite{MSCOCOVQA2015}.
In the experiments on these two data sets, our method outperforms the state of the arts in different well recognized metrics.

\subsection{Datasets}

\textbf{DAQUAR} contains 12,468 human question answer pairs on 1,449 images of indoor scene. The training set contains 795 images and 6,793 question answer pairs, and the testing set contains 654 images and 5,675 question answer pairs.

We run experiments for the full dataset with all classes, instead of their ``reduced set'' where the output space is restricted to only 37 object categories and 25 test images in total.
This is because the full dataset is much more challenging and the results are more meaningful in statistics.
The performance is reported using the ``Multiple Answers'' category but the answers are generated using open-ended approach.

\textbf{MSCOCO-VQA} is the latest VQA dataset that contains open-ended questions about images.
This dataset contains 369,861 questions and 3,698,610 ground truth answers based on 123,287 MSCOCO images. These questions and answers are sentence-based and open-ended.
The training and testing split follows MSCOCO-VQA official split. Specifically, we use 82,783 images for training and 40,504 validation images for testing.

\begin{figure*}[th]
  \centering
    \includegraphics[width=0.95\textwidth]{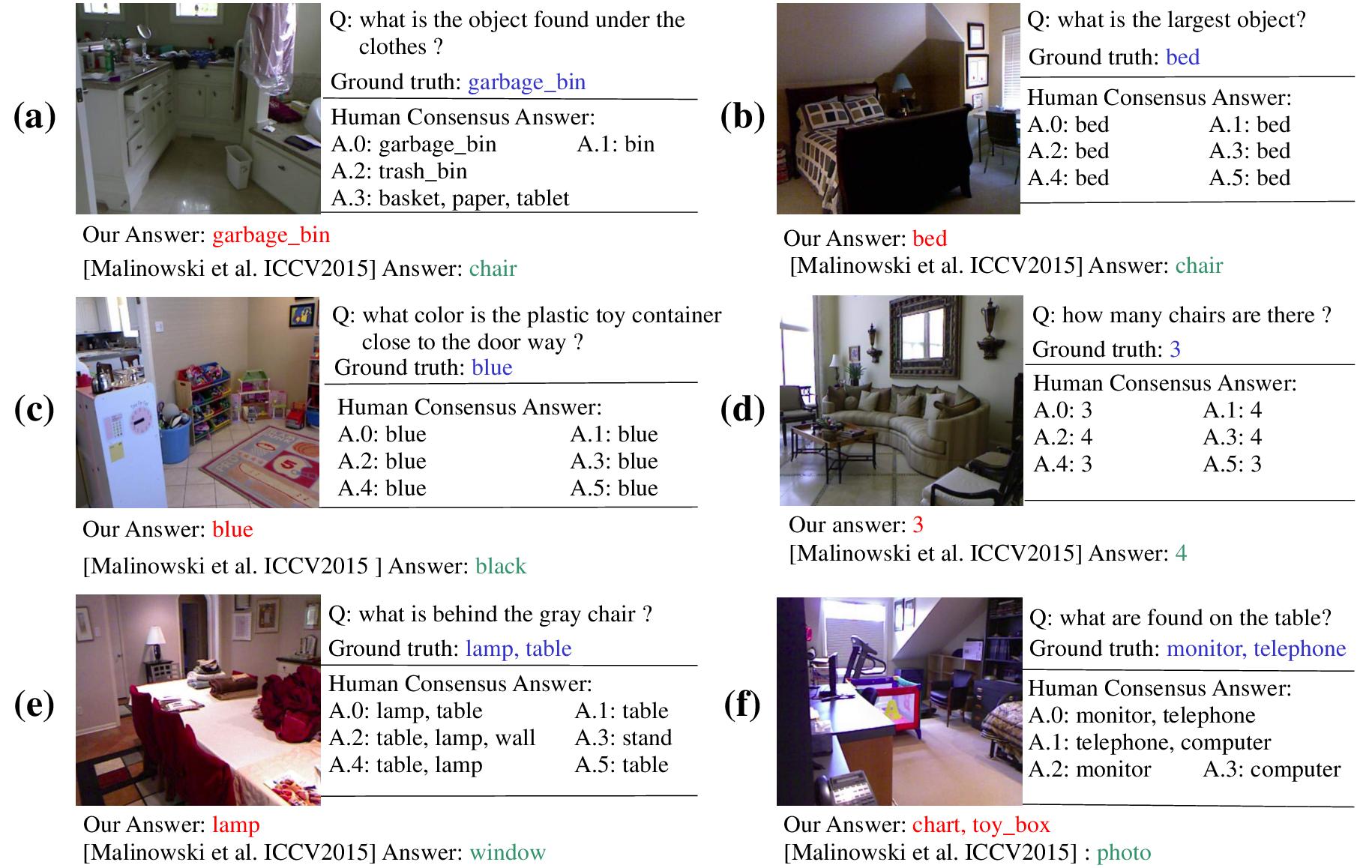}
  \caption{Examples of questions and answers on DAQUAR. Ground truth answers are colored in blue, our predicted answers in red and answers of state-of-the-art method in green. a)-d) are correct examples, and e)-f) are failure cases.}
  \label{fig:VQAexamples_DAQUAR}
\end{figure*}

\subsection{Evaluation criteria}

\subparagraph{DAQUAR} On the DARQUAR dataset, we  use the Wu-Palmer Similarity (WUPS) \cite{Wu:1994:VSL:981732.981751} score at different thresholds for comparison.
There are three metrics: Standard Metric, Average Consensus Metric and Min Consensus Metric. The Standard Metric is the basic score. The last two metrics are used to study the effects of consensus in question answering tasks.
Please refer to \cite{Malinowski_NIPS2014} for the details.

\subparagraph{MSCOCO-VQA} On the MSCOCO-VQA dataset, we use the evaluation criteria provided by the organizers.
For the open-ended tasks, the generated answers are evaluated using accuracy metric. It is computed as the percentage of answers that exactly agree with the ground truth provided by human. Please refer to \cite{MSCOCOVQA2015} for details.

\subsection{Experimental settings} \label{sec:setting}
We choose AlexNet and GoogLeNet in our experiments, respectively. 
For AlexNet, the region features are from Pool5 layer. For GoogLeNet we use the inception\_poo5b/output layer. 
That means our Compositional Memory processes flattened $36$ regions for AlexNet, and $49$ for GoogLeNet.

In our current implementation, the parameters of region-LSTMs and $\alpha$ gates in Compositional Memory are shared across regions, therefore the computational burden is minimal. However, as regions have to store their respective memory states, the storage space is more than traditional LSTM. In our experiments, the dimension of region inputs $d$ is 1024 for GoogLeNet and 256 for AlexNet. The dimension of the episodes is set to 200 for all LSTMs (including our Compositional Memory) on the DAQUAR dataset and the MSCOCO-VQA.

We used separate dictionaries for training and testing. To generate the dictionary, we first remove the punctuation marks, except those used in timing and measurements, separate them and convert them to lower cases.

We stop our training procedure after 100,000 iterations. The base learning rate is set to be 0.01. 
During training, it takes about 0.4 sec for one iteration on GTX Nvidia 780.

\subsection{Results on DAQUAR}

\subsubsection{Examples}
We first show some examples of the proposed method.
Figure~\ref{fig:VQAexamples_DAQUAR}a-\ref{fig:VQAexamples_DAQUAR}d show some correct examples and Figure~\ref{fig:VQAexamples_DAQUAR}e-\ref{fig:VQAexamples_DAQUAR}f are failure cases.
These examples demonstrate the effectiveness of the local features. 

When questions are sophisticated and location dependent, local features help most.
For example (Figure~\ref{fig:VQAexamples_DAQUAR}a and Figure~\ref{fig:VQAexamples_DAQUAR}c), the object name and the color cannot be easily obtained without the focus of the local region. 
Take Figure~\ref{fig:VQAexamples_DAQUAR}d) for another example, the number of chairs need to be counted. These cannot happen without the help of the local feature information encoded in the episodes.

We also show some failure cases in Figure~\ref{fig:VQAexamples_DAQUAR}e) and ~\ref{fig:VQAexamples_DAQUAR}f).
We observe that these are challenging cases, and even the human answers are not consistent.
Yet, our model is still able to find partially correct answers. 
For example (Figure~\ref{fig:VQAexamples_DAQUAR}e), while our answer does not fully match the ground truth, we provide the information that is much closer to the comparison methods.
It also should be noted that, although our answers may be incorrect (Figure~\ref{fig:VQAexamples_DAQUAR}f), we can still see the answers being related to object near the ground-truth objects.
These answers, albeit wrong, still show that our regional feature episodes are  useful and provide potentially meaningful answers.

\subsubsection{Comparisons with state-of-the-art methods}
We compare our proposed model with \cite{Malinowski-ICCV15,Malinowski_NIPS2014}.
Because they used different CNN methods, we tested both AlexNet and GoogLeNet. 
The performance on ``Multiple Answers'' (``Open-ended'') category are shown in Table~\ref{DAQUAR:Comparisons}.
\begin{table}[tbp]
\centering
\caption{Comparisons on full DAQUAR dataset, ``Multiple Answers'' category. The numbers are shown in percentage.}
\label{DAQUAR:Comparisons}
\begin{tabular}{|l|c|c|c|}
\hline
\multicolumn{1}{|c|}{\multirow{2}{*}{}} & \multirow{2}{*}{Accuracy} & \multicolumn{2}{c|}{WUPS}       \\ \cline{3-4}
\multicolumn{1}{|c|}{}                  &                           & @0.9           & @0.0           \\ \hline
\multicolumn{4}{|l|}{\textit{\textbf{Standard Metric}}}                                               \\ \hline
Malinowski et al. \cite{Malinowski_NIPS2014}       & 7.86                      & 11.86          & 38.79          \\ \hline
Ask-Neurons \cite{Malinowski-ICCV15} & 17.57                     & 23.31          & 57.49          \\ \hline
Our Model (AlexNet)                 & 21.92                     & 27.67          & \textbf{62.74} \\ \hline
Our Model (GoogleNet)               & \textbf{24.37}            & \textbf{29.77} & 62.73 \\ \hline
Human Answers                           & 50.20                     & 50.82          & 67.27          \\ \hline
\multicolumn{4}{|l|}{\textit{\textbf{Average Consensus Metric}}}                                      \\ \hline
Ask-Neurons \cite{Malinowski-ICCV15}          & 11.31                     & 18.62          & 53.21          \\ \hline
Our Model (AlexNet)                 & 14.72                     & 22.58          & \textbf{58.17} \\ \hline
Our Model (GoogleNet)               & \textbf{16.29}            & \textbf{23.95} & 57.68          \\ \hline
\multicolumn{4}{|l|}{\textit{\textbf{Min Consensus Metric}}}                                          \\ \hline
Ask-Neurons \cite{Malinowski-ICCV15}            & 22.74                     & 30.54          & 68.17          \\ \hline
Our Model (AlexNet)                 & 29.48                     & 37.60          & \textbf{75.16} \\ \hline
Our Model (GoogleNet)               & \textbf{31.52}            & \textbf{39.30} & 74.51          \\ \hline
\end{tabular}
\end{table}

The statistical results shows that the performance of our model substantially is better than the state of the art. On the WUPS$@$0.9, our method is $6\%$ higher (from $23.31\%$ to $29.77\%$).
When we lower the threshold in the WUPS, we are $5.25\%$ superior than the state of the art.

In other two measurements, where ``consensus'' is calculated on the answers from multiple subjects, our method also outperforms the state of the art $2\%$ to $7\%$.
This further confirms our system is more accurate and robust.

We also find that our method has better performance when GoogLeNet is used in our framework, although the difference is marginally noticeable.


\subsubsection{Effectiveness of Compositional Memory module}
We further present our study on the effectiveness of Compositional Memory and the Language LSTM in our VQA system.
Specifically, we show the comparison in performance when we toggle on and off these components. 

We consider the configuration where all modules are used as the ``full model'', and also name the configuration of traditional approach (Fig. \ref{fig:basicVQALSTM}) as ``baseline''. 
We then introduce three variants, where only language is used (``Factorized Language Only''), only Compositional Memory is used (``Episodes Only''), and both are used together (``Language+Episodes''). 



\begin{figure}[th]
  \centering
    \includegraphics[width=0.97\columnwidth]{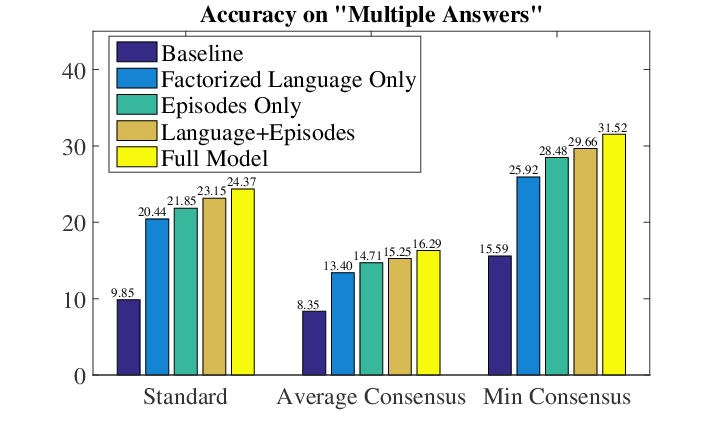}
    \caption{Comparisons of different variants of our model.}
  \label{fig:AccuraySelfComparison}
\end{figure}

The statistic results are shown in Figure~\ref{fig:AccuraySelfComparison} and Table~\ref{DAQUAR:selfcomparison_googlenet_standardmetric}. 
One can easily see that, the performance drops seriously when all the proposed memory are toggled off.
The episodes provide more information than the language, because it contains visual information from local image patch.
All together they achieve the score that each of them cannot reach.
This demonstrates that our Compositional Memory module contains critical information for visual question answering and the importance of the Language LSTM.

Using regional information does not rule out the importance of the visual context.
This holistic image features describes the abstraction of image.
One can see that the performance decreases, for example, from $24.37\%$ to $23.15\%$ on standard accuracy without the context.

Table~\ref{DAQUAR:selfcomparison_googlenet_standardmetric} shows the performance under different threshold in the WUPS metric.
For example, it is over 4\% better than the ``Language Only'' variant.
Since language semantics are important for answering questions and logical reasoning, while regional contents are more critical for answering questions about the existence of objects in image, Their fusion can further improve the qualities of answers.
As shown in ``Language+Episodes'', this fusion increases WUPS@0.9 from $25.77\%$ to $28.73\%$. 
With all components, our full model is consistently better than other variants.

As a conclusion, these three types of information are complementary and their combinations improve the solution of the VQA problem.
\begin{figure*}[th]
  \centering
    \includegraphics[height=0.18\textwidth]{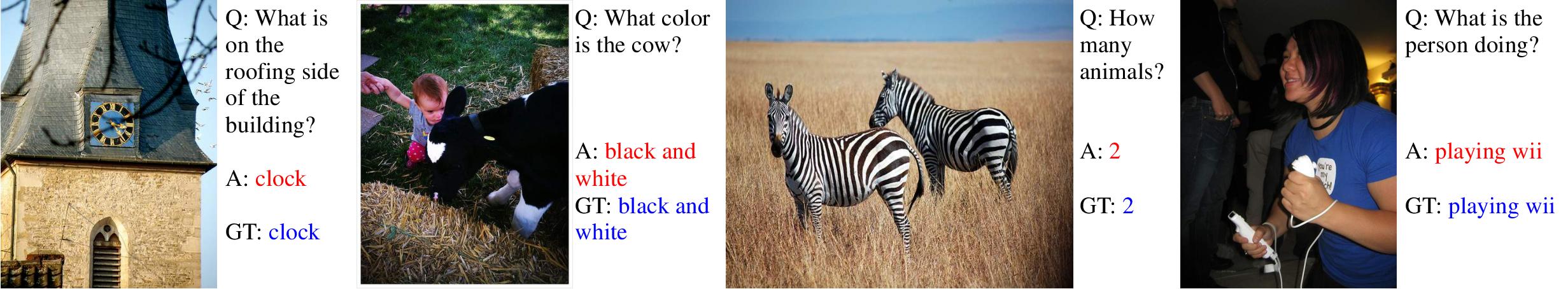}
  \caption{Examples on MSCOCO-VQA. Ground truth answers are colored in blue, and our predicted answers in red.}
  \label{fig:VQAexamples_VQA}
\end{figure*}

\begin{table}[tbp]
\small
\centering
\caption{WUPS@R: Comparisons of different variants of our model.}
\label{DAQUAR:selfcomparison_googlenet_standardmetric}
\begin{tabular}{ll|l|l|l|l|}
\hline
\multicolumn{2}{|c}{\multirow{2}{*}{}}                                                                              & \multicolumn{4}{|c|}{WUPS}                                                    \\ \cline{3-6}
\multicolumn{2}{|c}{}                                                                                               & \multicolumn{1}{|c|}{@0.9} & @0.7           & @0.5           & @0.0           \\ \hline
\multicolumn{2}{|l|}{Baseline} & 15.64                     & 36.12          & 52.43          & 54.41          \\ \hline

\multicolumn{2}{|l|}{Factorized Language Only}     & 25.77                     & 47.90          & 59.00          & 59.52          \\ \cline{3-6}
\hline
\multicolumn{2}{|l|}{Episodes Only}     & 27.43                     & 49.92          & 60.95          & 61.35          \\ \hline
\multicolumn{2}{|l|}{Language + Episodes}          & 28.73                     & 51.16          & 61.28          & 61.70          \\ \hline\multicolumn{2}{|l|}{\textbf{Full Model}}                                                                      & \textbf{29.77}            & \textbf{52.64} & \textbf{62.35} & \textbf{62.73} \\ \hline
\end{tabular}
\end{table}

\subsection{Results on MSCOCO-VQA}
Compared to DAQUAR, MSCOCO-VQA is the latest VQA dataset. It is  much larger and contains more scenes and question types that are not covered by DARQUAR.

Possibly because this is the latest outcome, there are different ways of evaluating the performances and reporting the results.
For example, while the measurement of accuracy is well defined, the evaluation protocols are not standardized.
Some practitioners use the organizer's previous release for training and validating, and further split the validation sets.
Only until recently the organizers release their \texttt{test-dev} set online, however, there are still many ways of handling the input. 
For example, the official version \cite{MSCOCOVQA2015} selects the most frequent $1000$ answers, which covers only $82.67\%$ of the answer set.
Different selections of dictionary can lead to fluctuations in the accuracy.
Finally, the tokenizers used in different practitioners may lead to other uncertainties in accuracy.

Due to the above concerns, we conclude that it is in the early stage of the evaluation, and would like to clearly outline our practices when readers examine the numbers.

\begin{itemize}
	\vspace{-8pt}
	\item We used a naive tokenizer as specified in Sec. \ref{sec:setting}.
	\vspace{-8pt}
    \item We used 13,880 words appeared in the training + validation answer set as our answer dictionary.
	\vspace{-8pt}
    \item We report results on both the \texttt{test-dev} and the full validation set.
\end{itemize}

We first show results of our method in Fig. ~\ref{fig:VQAexamples_VQA}.
Compared to Fig. \ref{fig:VQAexamples_DAQUAR}, one can see that MSCOCO-VQA is more diversified.
More results are shown in Supplementary Materials.

\subsubsection{Statistical results}
MSCOCO-VQA is grouped to a number of categories based on the types of the questions, and the types of answers.
We show the statistics on both categories in this section.

\subparagraph{Answer type} 
We report the overall accuracy and those of different answer types using both \texttt{text-dev} and the full validation set (Table \ref{VQA:per-answer-type-result}). 
Please note that we used a larger answer dictionary, which means potentially it is more difficult to deliver correct answers, but still our method achieved similar performance of the state of the art.

One can notice that the accuracy of simple answer type (\textit{e.g.} ``yes/no'') is very high, but the accuracies drop significantly when the answers become more sophisticated.
This indicates the potential issues and directions of our method.

\begin{figure}[th]
  \centering
    \includegraphics[width=1.05\columnwidth]{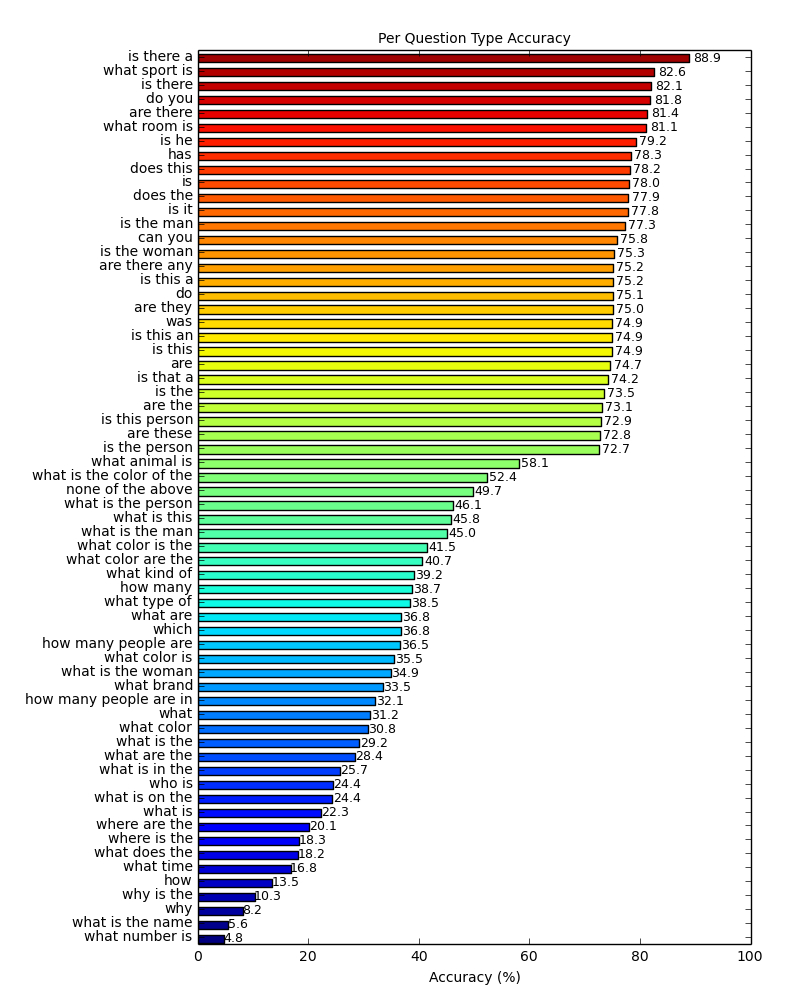}
  \caption{Accuracy on the Open-ended task of MSCOCO-VQA (validation) for different question types.}
  \label{fig:vqa_per_type_accuracy}
\end{figure}

\subparagraph{Question type}
We use validation set to report the accuracy of our method when question type varies (Figure~\ref{fig:vqa_per_type_accuracy}).
The colored bar chart and sorted according to the accuracy, with the numbers displayed next to the bars.

It is interesting to see a significant drop when the questions ascend from simple forms (\textit{e.g.}, ``is there?'') to complicated ones (\textit{e.g.}, ``what'',``how'').
This suggest that a practical VQA system needs to take this prior into consideration.

\begin{table}[h]
\centering
\caption{Results on MSCOCO-VQA (Open-ended). Red, blue and green are used to denote top 3. The baseline is \textbf{LSTM Q+I} in \cite{MSCOCOVQA2015}.}
\label{VQA:per-answer-type-result}
\begin{tabular}{|l|c|c|c|c|}
\hline
              & All  & Yes/No  & Number & Other \\
\hline
Baseline & {\color{red} 53.74}  &  {\color{blue}78.94}  & {\color{blue}35.24} & {\color{red}36.42} \\
\hline
Ours (test-dev) &  {\color{blue}52.62}  & {\color{green}78.33} & {\color{red}35.93} & {\color{blue}34.46} \\ 
\hline
Ours (val) &  {\color{green}50.48}  & {\color{red}79.05} & {\color{green}32.60} & {\color{green}33.59} \\ 
\hline
\end{tabular}
\end{table}




\section{Conclusion}
In this paper we propose to use the Compositional Memory as the core element in the VQA.
Our end-to-end approach is capable of dynamically extracting local the features.
The Long Short-Term Memory (LSTM) based approach fuses image regions and language, and generates the episodes that is effective for high level reasoning.
Our experiments on the latest public datasets suggest that our method has a superior performance.

{\small \bibliographystyle{ieeetr}
\bibliography{CVPR-2016-VQA}
 }
\end{document}